\documentclass[letterpaper]{article} 
\usepackage[draft]{aaai25}  
\usepackage{times}  
\usepackage{helvet}  
\usepackage{courier}  
\usepackage[hyphens]{url}  
\usepackage{graphicx} 
\usepackage{natbib}  
\usepackage{makecell}
\usepackage{caption} 
\usepackage{algorithm}
\usepackage{algorithmic}

\usepackage{url}            
\usepackage{booktabs}       
\usepackage{amsfonts}       
\usepackage{nicefrac}       
\usepackage{microtype}      
\usepackage{xcolor}         
\usepackage{graphicx}
\usepackage{multirow,tabularx}
\usepackage{amsmath,amsfonts}
\usepackage{subcaption}
\usepackage{arydshln}
\usepackage{soul}

\usepackage{newfloat}
\usepackage{listings}

\DeclareCaptionStyle{ruled}{labelfont=normalfont,labelsep=colon,strut=off} 
\floatstyle{ruled}
\newfloat{listing}{tb}{lst}{}
\floatname{listing}{Listing}
\title{Enhancing Fine-grained Image Classification through Attentive Batch Training}
\author{
    Duy M. Le\textsuperscript{\rm 1}\equalcontrib, Bao Q. Bui\textsuperscript{\rm 2}\equalcontrib, Anh Tran\textsuperscript{\rm 3}, Cong Tran\textsuperscript{\rm 1}, Cuong Pham\textsuperscript{\rm 1}
}
\affiliations{
    \textsuperscript{\rm 1}Post \& Telecommunication Institute of Technology\\
    \textsuperscript{\rm 2}VinUniversity\\
    \textsuperscript{\rm 3}VinAI Research

}

\begin{document}

\maketitle

\begin{abstract}
Fine-grained image classification, which is a challenging task in computer vision, requires precise differentiation among visually similar object categories. In this paper, we propose 1) a novel module called Residual Relationship Attention (RRA) that leverages the relationships between images within each training batch to effectively integrate visual feature vectors of batch images and 2) a novel technique called Relationship Position Encoding (RPE), which encodes the positions of relationships between original images in a batch and effectively preserves the relationship information between images within the batch.
Additionally, we design a novel framework, namely Relationship Batch Integration (RBI), which utilizes RRA in conjunction with RPE, allowing the discernment of vital visual features that may remain elusive when examining a singular image representative of a particular class. 
Through extensive experiments, our proposed method demonstrates significant improvements in the accuracy of different fine-grained classifiers, with an average increase of $(+2.78\%)$ and $(+3.83\%)$ on the CUB200-2011 and Stanford Dog datasets, respectively, while achieving a state-of-the-art results $(95.79\%)$ on the Stanford Dog dataset. Despite not achieving the same level of improvement as in fine-grained image classification, our method still demonstrates its prowess in leveraging general image classification by attaining a state-of-the-art result of $(93.71\%)$ on the Tiny-Imagenet dataset. Furthermore, our method serves as a plug-in refinement module and can be easily integrated into different networks.
\end{abstract}

%

\section{Introduction}
Fine-grained classification is an important task in computer vision as it has a wide range of real-world applications, including image recognition, disease diagnosis \cite{disease1,disease2,disease3}, or  biodiversity monitoring \cite{bio1,bio2,bio3}, where distinguishing between visually similar subcategories is crucial. With the rapid advancement of technology, we now have the capability to collect and store a large amount of image data from various sources. 
Nevertheless, the classification of objects in images characterized by a high degree of similarity, commonly referred to as the {\em fine-grained image classification} problem—examples include the categorization of bird species, types of leaves, or models of electronic products—presents a noteworthy and persistent challenge.

In addition to image classification in general, fine-grained image classification exposes more significantly challenging, including (i) substantial intra-class variation, with objects in the same category exhibiting significant pose and viewpoint differences; (ii) subtle inter-class distinctions, where objects from different categories may closely resemble each other with minor differences; (iii) constraints on training data, as labeling fine-grained categories often demands specialized expertise and a substantial amount of annotation effort \cite{He_2017}. For these reasons, fine-grained classification remains an open topic for research community. 

\begin{figure}
\includegraphics[width=\linewidth]{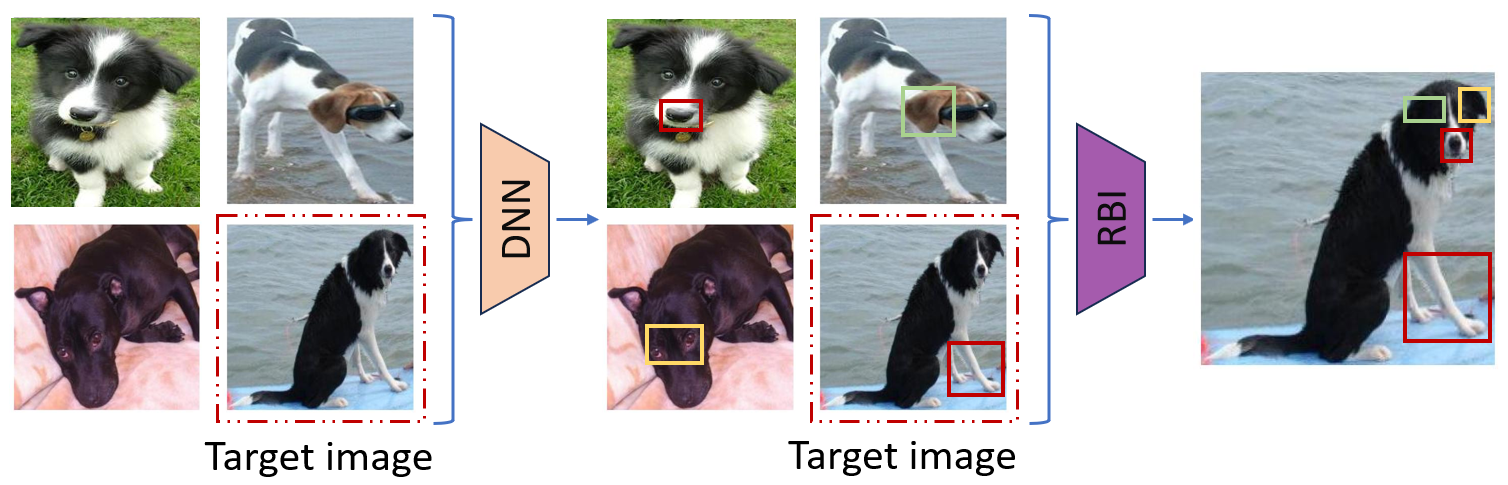}
\caption{Example of intra-batch feature fusion to enhance predictivity for target images.}
\label{fig:motivation}
\vspace{-0.5cm}
\end{figure}
Motivated by the fact that deep neural networks (DNNs) encounter challenges in effectively distinguishing intricate features and grappling with the inherent complexities of learning detailed patterns, our study centers on scrutinizing the mechanisms by which image feature extractors discern nuanced features. Additionally, our investigation delves into the process of {\em explicitly} amalgamating these discerned features across a batch of images to construct intricate feature maps, ultimately enhancing the accuracy of fine-grained image classification.
Figure \ref{fig:motivation} illustrates an example of our idea, integrating subtle features extracted from several images to generate a sophisticated feature. In the figure, the delineated bounding boxes within each image delineate regions of interest identified by the model for focused attention. Boxes sharing the same color signify images belonging to identical classes, while distinct colors indicate diverse classification categories. We aim at discerning specific connections between the reference image and others, facilitating the amalgamation of subtle features derived from various images, even those belonging to different categories, to complement the features of the target image. Concretely, as exemplified in the provided figure, features corresponding to the snout, ear, and eye regions extracted from three distinct images are harmonized with the feet region. This integration results in a cohesive feature representation that concentrates on multiple salient positions of the dog within the target image. Intuitively, we can design a novel framework for this purpose, termed as Relationship Batch Integration (RBI).

Our contributions can be summarized in the following:
\begin{itemize}
    \item {We present a novel attention module, termed {R}esidual {R}elationship {A}ttention (RRA), that effectively exploits subtle features from multiple images and combines them to generate sophisticated features, particularly in the case of a CNN-based backbone. Additionally, it improves the focusing ability on important features when utilizing an Attention-based backbone. To this end, it addresses the issue of ambiguity in fine-grained classification, mitigating its impact.}
    \item {We introduce a novel technique called Relationship Position Encoding (RPE), which effectively preserves the relationship information between images within a batch.}
    \item {We introduce a novel RBI framework that integrates RRA and RPE, offering seamless integration into fine-grained classifiers. Remarkably, even when employing a reduced version of the pre-trained backbone  with fewer parameters compared to the larger baselines, the RBI improves both accuracy and processing time. }
    \item Our extensive experiments on publicly available datasets demonstrate the model's capability to enhance feature clustering and accuracy, while also achieving state-of-the-art results on the CUB200-2011 and Stanford Dogs datasets.
\end{itemize}


\section{Related work}

\label{related_work}
This section investigates the previous works in attention-based methods and fine-grained image classification.


\subsection{Attention-based methods}
\noindent{\textbf{Attention-based methods}} have gained significant attention and achieved notable success in various domains of machine learning and computer vision. These methods enable models to focus on specific parts or features of input data, allowing for more effective utilization of information. One seminal work in this area is the Transformer model \cite{vaswani2017attention}. The Transformer model utilizes self-attention mechanisms to capture dependencies between different positions in an input sequence, facilitating the modeling of long-range dependencies. 

\noindent{\textbf{Computer vision}}. Attention mechanisms have been widely adopted to improve visual understanding tasks. Xu \textit{et al.} \cite{xu2015show} proposed the spatial attention mechanism in the context of image captioning, enabling models to selectively attend to relevant regions of an image when generating descriptions. Other works, such as Hu \textit{et al.} \cite{hu2018squeeze} and Woo \textit{et al.} \cite{woo2018cbam}, introduced attention mechanisms in convolutional neural networks (CNNs) to enhance feature representation and capture fine-grained details. These attention-based CNNs have achieved state-of-the-art results in image classification and object detection.

\noindent{\textbf{Reinforcement learning}}. Mnih\textit{ et al.} \cite{mnih2014recurrent} introduced the attention-based model called the Recurrent Attention Model (RAM), which learns to sequentially attend to different regions of an image to perform tasks such as visual attention and object recognition.

\subsection{Fine-grained image classification}
Recent deep learning research on fine-grained classification has primarily focused on two main directions:  convolutional neural networks (CNN)-based and visual attention-based methods.

\noindent{\textbf{CNN-based Fine-Grained Image Classification} is commonly seen in general classification tasks and specifically in fine-grained classification problems. Common baseline CNN architectures such as MobileNet \cite{mobilenets}, DenseNet \cite{densenet}, ConvNeXt \cite{convnext}, and others can also be applied to fine-grained classification tasks. 
Furthermore, there exist unique approaches, such as implicitly separating the class-relevant foreground from the class-irrelevant background \cite{journal1}, that aim to enhance the performance of the model. In a notable achievement, the CNN-based P2P-Net \cite{P2P-Net} model, published at CVPR 2022, achieved top performance accuracy on the CUB-200-2011 datasets.

\noindent\textbf{Visual attention-based approaches
} aim to mimic human visual attention by selectively focusing on informative regions or features within an image. One of the pioneering models utilizing this mechanism, \cite{2:238}, uses two-level attention to concentrate on both overall image context and fine-grained details. More recently,  a reinforcement learning-based fully convolutional attention localization network \cite{2:241} adaptively selects multiple task-driven visual attention regions. This model is renowned for being significantly more computationally effective in both the training and testing phases. Furthermore, the ViT-NeT \cite{ViT-NeT} model augments the explicability of Vision Transformers \cite{ViT} by integrating a neural tree decoder, enabling the generation of predictions with hierarchical structures that facilitate improved comprehension and examination of the model's decision-making process. In another context, HERBS \cite{HERBS} employs two innovative approaches, namely high-temperature refinement and background suppression, to address key challenges in fine-grained classification.
Notably, the Attention-based TransFG \cite{TransFG} model demonstrated exceptional accuracy in fine-grained image classification datasets, attaining top performance.
Currently, the ViT-NeT and HERB models achieved the highest accuracies on the Stanford Dogs dataset \cite{datasetFGVC2011}, and the NABirds dataset \cite{NABirds}, respectively. 

\section{Proposed Approach}
\label{sec:proposed_approach}


\subsection{Relationship Batch Integration (RBI) Framework}

As illustrated in Figure \ref{fig:figure1}, the Relationship Batch Integration (RBI) Framework is composed of three components: \textit{relationship position encoding (RPE)},\textit{ deep neural network (DNN) feature extractor}, and \textit{residual relationship attention (RRA) module}. Specifically, for a given batch of input images $\mathbf{X} \in \mathbb{R}^{B \times 3 \times H \times W}$ ($B$ denote the number of images within the batch, while $H$ and $W$ represent the height and width of an image in $\mathbf{X}$, respectively), we begin by calculating the similarity between each pair of input images using the relationship position encoding module, described in details in Section \ref{RPE} 
. This process produces a 
{similarity} matrix $\mathbf{S} \in \mathbb{R}^{B \times B}$. Simultaneously, $\mathbf{X}$ is passed through the DNN feature extractor, which can be a CNN-based backbone such as Densenet \cite{densenet}, ConvNeXt \cite{convnext}, or an Attention-based backbone such as Swin Transformer \cite{swinT}, ViT \cite{ViT}. This yields DNN feature embeddings $\mathbf{N} \in \mathbb{R}^{B \times D}$ (where $D$ signifies the dimensionality of the DNN feature embeddings). Subsequently, the similarity matrix $\mathbf{S}$ and DNN feature embeddings $\mathbf{N}$ are utilized as inputs for the RRA module. Finally, the output of the RRA module is forwarded to fully connected layers to generate logits for classification.

\begin{figure*}
  \centering
   \includegraphics[width=.95\textwidth]{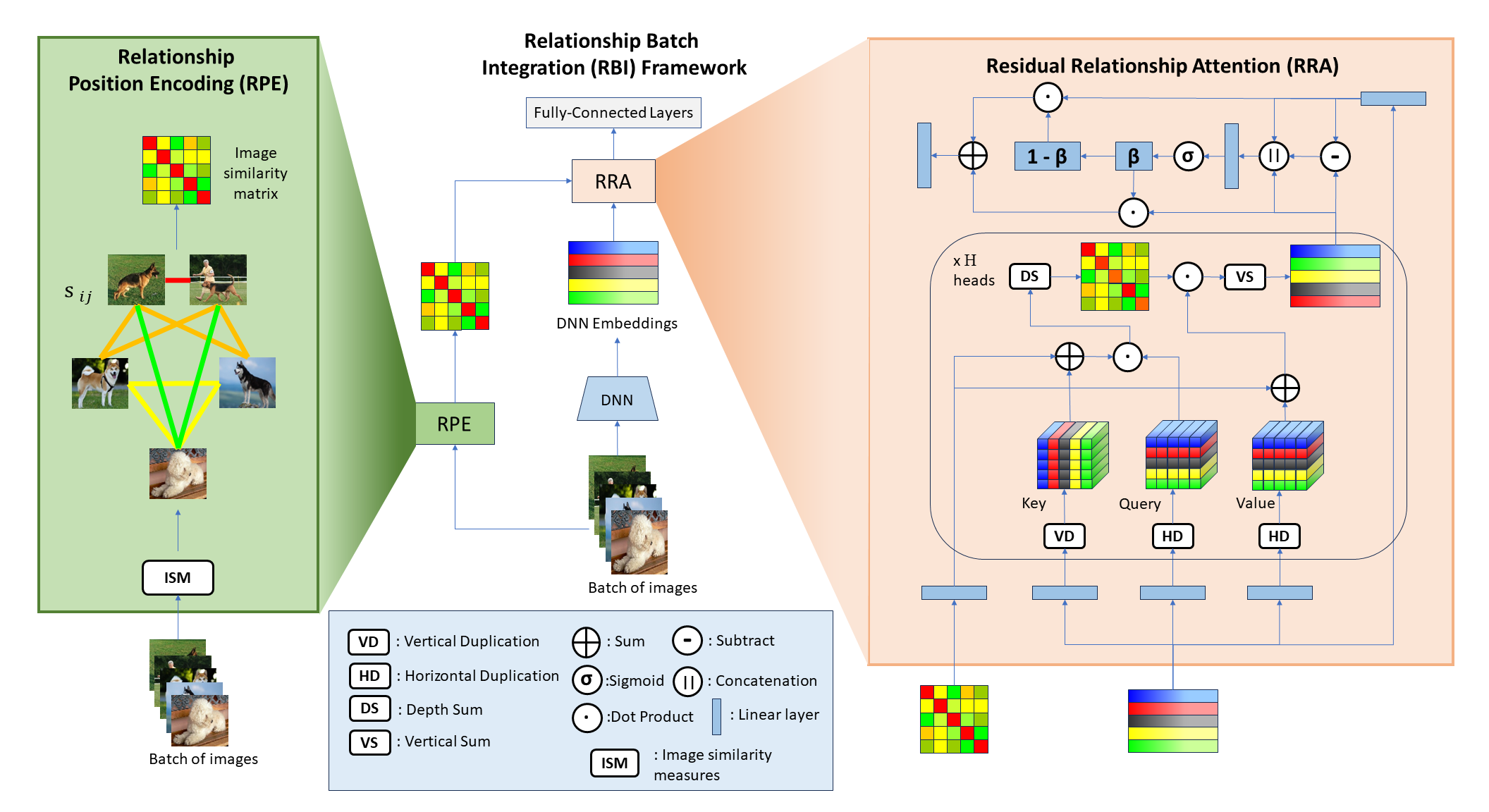}   
   \caption{Relationship Batch Integration (RBI) Framework}
   \label{fig:figure1}
   \vspace{-0.3cm}
\end{figure*}

\subsection{Relationship Position Encoding (RPE)}
\label{RPE}

Contrary to prior studies such as \cite{vaswani2017attention}, which commonly utilize position encoding to incorporate information about the relative or absolute position of tokens in a sequence, the order of images in the batch is not essential in this case. This is attributed to the random sampling of images from the dataset, devoid of any inherent relational structure. However, it is important to preserve the position of relationships between images within the batch during the integration process within the RRA module. Thus, we devise a Relationship Position Encoding to serve the crucial purpose of embedding this positional insight into the RRA module.

{In this study, the {similarity} {weight $s_{ij}$ ($s_{ij} \in \mathbf{S}$)} 
between two images is determined by RPE. To begin, we compute the Mean Squared Error (MSE) between two images  \( \mathbf{I}_i \) and \( \mathbf{I}_j \) within a batch (where \( \mathbf{I}_i, \mathbf{I}_j \in \mathbb{R}^{3 \times H \times W} \)):
\begin{equation}
\operatorname{MSE}(\mathbf{I}_i, \mathbf{I}_j) = \frac{1}{3HW} \sum_{x=0}^{2} \sum_{y=0}^{H-1} \sum_{z=0}^{W-1} \left[\mathbf{I}_i(x, y, z) - \mathbf{I}_j(x, y, z)\right]^2.
\end{equation}
Then, we compute $s_{ij}$ in the form of a normalized PSNR by  incorporating a very small coefficient \(\epsilon\)  to prevent division by zero when the two images are identical:
\begin{equation}
s_{ij} = \widehat{\operatorname{PSNR}}(\mathbf{I}_i, \mathbf{I}_j) = 20 \cdot \log_{10}\left(\frac{\operatorname{MAX}_{\mathbf{I}_i}}{\sqrt{\operatorname{MSE}(\mathbf{I}_i, \mathbf{I}_j)} + \epsilon}\right);
\label{eq:psnr}
\end{equation}
where \(\operatorname{MAX}_{\mathbf{I}_i}\) represents the maximum possible pixel value of \(\mathbf{I}_i\).} 


\subsection{Residual Relationship Attention (RRA)}

Before delving deeper into the RRA structure, we first establish several definitions for mathematical operations.

\hspace{1mm}\textbf{Vertical Duplication.} Given DNN feature embeddings $\mathbf{N} \in \mathbb{R}^{B \times D}$ as input. Initially, the operation involves expanding the tensor along its first dimension, transforming it from $\mathbf{N} \in \mathbb{R}^{B \times D}$ to $\mathbf{N} \in \mathbb{R}^{1 \times B \times D}$. Subsequently, the vertical duplication operation duplicates each instance along the first dimension. It concatenates them along the same dimension, resulting in a tensor of shape $\mathbb{R}^{B \times B \times D}$, formulated as:
\begin{equation}
    \mathcal{D}_v(\mathbf{N}) = \parallel_{i=1}^{B} \mathbf{N} \in \mathbb{R}^{1 \times B \times D} \rightarrow \mathbb{R}^{B \times B \times D}.
\end{equation}

\textbf{Horizontal Duplication.} 
The horizontal duplication operation involves unsqueezing the second dimension of the input and concatenating the duplicated inputs along the same dimension:
\begin{equation}
    \mathcal{D}_h(\mathbf{N}) = \parallel_{i=1}^{B} \mathbf{N} \in \mathbb{R}^{B \times 1 \times D} \rightarrow \mathbb{R}^{B \times B \times D}.
\end{equation}

\textbf{Depth Sum.} The operation of depth sum involves the reduction of the input matrix $\mathbf{F}$, which is of dimension $B \times B \times D$, through summation along its last dimension. Mathematically, it can be expressed as:
\begin{equation}
    \mathcal{S}_d(\mathbf{F}) = \left[\sum_{k=1}^{D} f_{ijk}\right]_{i,j = \overline{1,B}} \in \mathbb{R}^{B \times B};
\end{equation}
where $f_{ijk}$ represents the value of the element at position $i, j, k$ in the matrix $\mathbf{F}$.

\hspace{1mm}\textbf{Vertical Sum.} Similarly, the vertical sum operation entails reducing the first dimension of the input matrix $\mathbf{F}$ through summation. It is represented as:
\begin{equation}
    \mathcal{S}_v(\mathbf{F}) = \left[\sum_{i=1}^{B} f_{ijk}\right]_{j = \overline{1,B}, k = \overline{1,D}} \in \mathbb{R}^{B \times D}.
\end{equation}

{Next, we proceed to explore the individual components of the RRA structure in greater detail.}

\hspace{1mm}\textbf{Keys.} The keys in the RRA are derived from the parameter matrix of keys, denoted as $\mathbf{W}_K \in \mathbb{R}^{D \times D}$, and DNN feature embeddings $\mathbf{N} \in \mathbb{R}^{B \times D}$:
\begin{equation}
    \mathbf{K} = \mathcal{D}_v(\mathbf{N} \mathbf{W}_K) \in \mathbb{R}^{B \times B \times D}.
\end{equation}

\textbf{Queries, Values.} Given weight matrices $\mathbf{W}_Q \in \mathbb{R}^{D \times D}$ and $\mathbf{W}_V \in \mathbb{R}^{D \times D}$ representing learnable parameters of queries and values, respectively, and DNN feature embeddings $\mathbf{N} \in \mathbb{R}^{B \times D}$, the queries $\mathbf{Q}$ and the values $\mathbf{V}$ in the RRA are calculated according to the following equations:
\begin{equation}
    \mathbf{Q} = \mathcal{D}_h(\mathbf{N} \mathbf{W}_Q) \in \mathbb{R}^{B \times B \times D};
\end{equation}
\begin{equation}
    \mathbf{V} = \mathcal{D}_h(\mathbf{N} \mathbf{W}_V) \in \mathbb{R}^{B \times B \times D}.
\end{equation}

\textbf{Attention.} To compute the attention matrix $\mathbf{A}$, given the relationship position embeddings $\mathbf{S} \in \mathbb{R}^{B \times B}$, queries $\mathbf{Q}$, and keys $\mathbf{K}$, the following equation is utilized:
\begin{equation}
    \mathbf{A} = \operatorname{softmax}\left(\mathcal{S}_d \left(\frac{\mathbf{Q} \cdot (\mathbf{K} + \mathbf{S})}{\sqrt{D}}\right)\right) \in \mathbb{R}^{B \times B}.
\end{equation}

\textbf{Attention Embeddings.} The attention embeddings $\mathbf{Z}$ are computed using the attention matrix $\mathbf{A}$, values $\mathbf{V}$, and relationship position embeddings $\mathbf{S}$:
\begin{equation}
    \mathbf{Z} = \mathcal{S}_v\big(\mathbf{A} \cdot (\mathbf{V} + \mathbf{S})\big) \in \mathbb{R}^{B \times D}.
\end{equation}

\textbf{Output.} Finally, the output $\mathbf{C}$ of the RRA is computed using attention embeddings $\mathbf{Z}$, parameter matrix of residual $\mathbf{W}_S \in \mathbb{R}^{D \times D}$, and DNN feature embeddings $\mathbf{N}$:
\begin{equation}
    \beta = \operatorname{sigmoid}\Big(\mathbf{W}_{\beta}\big[\mathbf{Z}||(\mathbf{W}_S \mathbf{N}) || (\mathbf{Z} - \mathbf{W}_S \mathbf{N})\big]\Big);
\end{equation}
\begin{equation}
    \mathbf{C} = \operatorname{BatchNorm}\big((1 - \beta)\mathbf{Z} + \beta \mathbf{W}_S \mathbf{N}\big);
\end{equation}
where $||$ represents the concatenation operation along the last dimension of tensors and $\mathbf{W}_{\beta}$ denotes the parameter matrix of beta.
Equations (12) and (13) facilitate the preservation of the original features of the DNN and enable flexible adjustment of the weights assigned to both the DNN and RRA features.
 

\section{Experiments}
\label{experiments}

\subsection{Datasets and Experimental Settings}
\label{sec:configurations}

\noindent\textbf{Datasets.} We perform experiments on three well-known fine-grained datasets: CUB-200-201 \cite{CUB-200-2011}, Stanford Dogs \cite{datasetFGVC2011}, and NABirds \cite{NABirds}.
\footnote{For a more comprehensive understanding of the detailed information about the datasets, please refer to Section Detailed Configuration of Datasets in Supplementary Material.}


\noindent\textbf{Implementation details.} All experiments are conducted on an NVIDIA Tesla T4 GPU with 15GB of RAM. Initially, all input images are resized to 224x224 pixels. We employ simple data augmentation techniques such as RandomHorizontalFlip and RandomRotation during training. 
The DNN encoder is trained using pre-trained weights from the ImageNet1K dataset. 
The model is fine-tuned for 50 epochs using a batch size of 32 for all models. As the proposed RBI can be influenced by the batch size, we provide detailed experiments to evaluate the results corresponding to different batch size configurations in section \ref{sec423}. We train the network using the Rectified Adam optimizer with a default epsilon value of $e^{-8}$. The dimension of the embedding of the encoder network is set to 1024. We evaluate the top-1 classification error on the shuffled validation set. Additionally, the initial learning rate is set to $e^{-5}$. \footnote{The source code of the implementation is available online (currently omitted due to blind review).}

\subsection{Comparison to Existing Methods}
\label{sec:comparison_accuracies}

\noindent\textbf{Baselines}. To validate the effectiveness and generalization of our method, we investigate the performance of incorporating RBI on four different well-known DNNs and their variants, including DenseNet \cite{densenet}, MobileNet \cite{mobilenets}, ConvNeXt \cite{convnext}, SwinTransformer \cite{swinT}, and HERB \cite{HERBS}. It is important to highlight that our RBI is the only modification, while all other training configurations and hyperparameters remain unaltered from the original implementations. Even though we incorporate our proposed method across various techniques and assess it on diverse datasets, we maintain the consistent parameter configuration detailed earlier throughout all experiments.

\begin{table*}[t]
\caption{The impact of RBI on fine-grained classification outcomes when incorporated into various DNN techniques. The accuracy gain when applying RBI is provided in the brackets. The bolded models highlight how the RBI model with a smaller pretrained backbone outperforms the baseline large model in both inference time and parameter count.}
\label{tab:compare_baseline}
\begin{center}
\centering
\begin{tabular}{l|c|c|c|c|c}
\hline 
\multirow{3}{*}{ Method } & \multirow{3}{*}{$\begin{array}{c}{\text{Inference}} \\
\text{time}\end{array}$} & \multirow{3}{*}{$\begin{array}{c}{\text{\# params}} \\
\text{}\end{array}$} & \multicolumn{3}{c}{ Acc (\%) } \\
\cline { 4 - 6 } & & & $\begin{array}{c}{\text{Stanford}} \\
\text{Dogs}\end{array}$ & $\begin{array}{c}\text{CUB-200-} \\
\text{2011}\end{array}$ & $\begin{array}{c}\text{NABirds} \\
\text{ }\end{array}$ \\
\hline MobilenetV3-S &0.013 &1.6M & 73.12 & 67.5 &  
66.46 \\
MobilenetV3-S-RBI &0.016 &17.4M & $77.01(+3.89)$ & $69.86(+2.36)$ & $69.1(+2.64)$  \\
 MobilenetV3-L &0.035 &4.4M & 78.31 & 77.65 &  75.86 \\
MobilenetV3-L-RBI &0.039 &23.2M & $82.72(+4.41)$ & $80.77(+3.12)$ &  $79.82(+3.96)$\\
\hline Densenet201 &0.28 &18.3M & 83.95 & 79.13 & 77.55 \\
\hdashline
Densenet201-RBI &0.29 &73.7M & $87.72(+3.77)$ & $84.48(+5.35)$ & $83.81(+6.26)$ \\
Densenet161 &0.42 &26.7M & 84.46 & 79.68 & 78.97 \\
\hdashline
Densenet161-RBI &0.45 &88.7M & $88.47(+4.01)$ & $84.79(+5.11)$ & $84.75(+5.78)$ \\
\hline SwinT-Small &0.51 &49.1M & 91.39 & 86.27 & 86.74 \\
\hdashline
\textbf{SwinT-Small-RBI} &0.52 &61.7M & $92.79(+1.40)$ & $87.35(+1.08)$ & $87.97(+1.23)$ \\
\textbf{SwinT-Big} &0.82 &87.0M & 92.11 & 85.86 & 86.32 \\
\hdashline
SwinT-Big-RBI &0.84 &102.8M & $93.06(+0.95)$ & $87.9(+2.04)$ & $88.03(+1.71)$ \\
\hline ConvNeXtBase &0.59 &88.7M & 92.77 & 81.93 & 85.31 \\
\hdashline
\textbf{ConvNeXtBase-RBI} &0.61 &103.4M & $94.56(+1.79)$ & $87.52(+5.59)$ & $87.86(+2.55)$ \\
\textbf{ConvNeXtLarge} &1.22 &197.9M & 93.71 & 81.74 & 85.53 \\
\hdashline
ConvNeXtLarge-RBI &1.23 &231.8M & $95.79(+2.08)$ & $87.8(+6.06)$ & $88.11(+2.58)$ \\
\hline HERB-SwinT &1.74 &286.6M & 88.62 & 89.9 & 90 \\
HERB-SwinT-RBI &1.88 &318.2M & $88.9(+0.28)$ & $90.37(+0.47)$ & $90.61(+0.61)$ \\
\hline TransFG & 0.85 &86.3M & 89.18 & 90.19 & 89.9 \\
TransFG-RBI &0.86 &95.2M & $89.76(+0.58)$ & $90.66(+0.47)$ & $90.45(+0.55)$ \\
\hline P2P-Net & 0.46 &63.4M & 83.14 & 86.07 & 85.12 \\
P2P-Net-RBI & 0.48 & 110.6M & $89.3(+6.16)$ & $91.07(+5.00)$ & $90.56(+5.44)$ \\
\hline \hline \textbf{Avg. Improvement} & & & \textbf{+2.51} & \textbf{+3.46} &  \textbf{+3.04} \\
\hline
\end{tabular}
\end{center}
\end{table*}

\noindent\textbf{Comparison results.} Table \ref{tab:compare_baseline} shows the impact of our RBI on fine-grained classification performance across different methods and datasets. Our interesting findings are summarized as follows:
\begin{itemize}
\item The table clearly illustrates that the incorporation of RBI consistently improves fine-grained classification results. Notably, we observe an average increase of $+2.78\%$, $+3.83\%$, and $+3.29\%$ on the Stanford Dogs, CUB-200-2011 datasets and NABirds, respectively. 
\item While RBI significantly enhances the performance of CNN-based models on both datasets, the improvement is more moderate for transformer-based models. We hypothesize that because of the inherent similarity between the attention mechanism of transformers and the nature of RRA, the accuracy improvement is not as substantial as with CNN-based models. For example, with models like DenseNet and MobileNet, accuracy increases by $3-6\%$ on both datasets, while with Swin Transformer, it ranges from $1-2\%$. Notably, ConvNeXt shows a slight performance boost on the Stanford Dogs dataset but a significant improvement of $5-6\%$ on CUB-200-2011.
\item Improving existing fine-grained classification methods is a challenging endeavor. However, as shown in Table \ref{tab:compare_baseline}, our proposed approach achieves new state-of-the-art results on the Stanford Dogs dataset\footnote{According to the comparison table in \url{https://paperswithcode.com/sota/fine-grained-image-classification-on-stanford-1} on 01/02/2024.}. 
It's crucial to note that the reported accuracies on the CUB-200-2011 and NABirds datasets diverge from those in the literature, such as HERB, TransFG, and P2P-Net, due to our uniform training configurations outlined in Section \ref{sec:configurations}, which may differ from the original methodologies.


\item Additionally, we observe that for some models, when we add the RBI module to smaller variants, they achieve better accuracy than the larger variants without the module, while also being less time-consuming and complex. For instance, SwinT-Small-RBI (61.7M parameters) outperforms SwinT-Big (87M parameters), and ConvNeXtBase-RBI (103.4M parameters) surpasses ConvNeXtLarge (197.9M). This partly demonstrates the effectiveness of the proposed module when integrated into different backbones. Regarding neural network complexity, despite a significant increase in the number of parameters in the proposed models compared to the base ones, the inference time varies only slightly between them.
\item Furthermore, our observations reveal that the inclusion of RBI in all smaller model versions outperforms their larger counterparts in terms of inference time. This comparison is fairer than solely considering the number of parameters, especially for compact models like MobileNet and DenseNet, which exhibit a remarkable speed improvement of up to twice as fast.
\end{itemize}


\begin{figure}
  \centering
   \includegraphics[width=\linewidth]{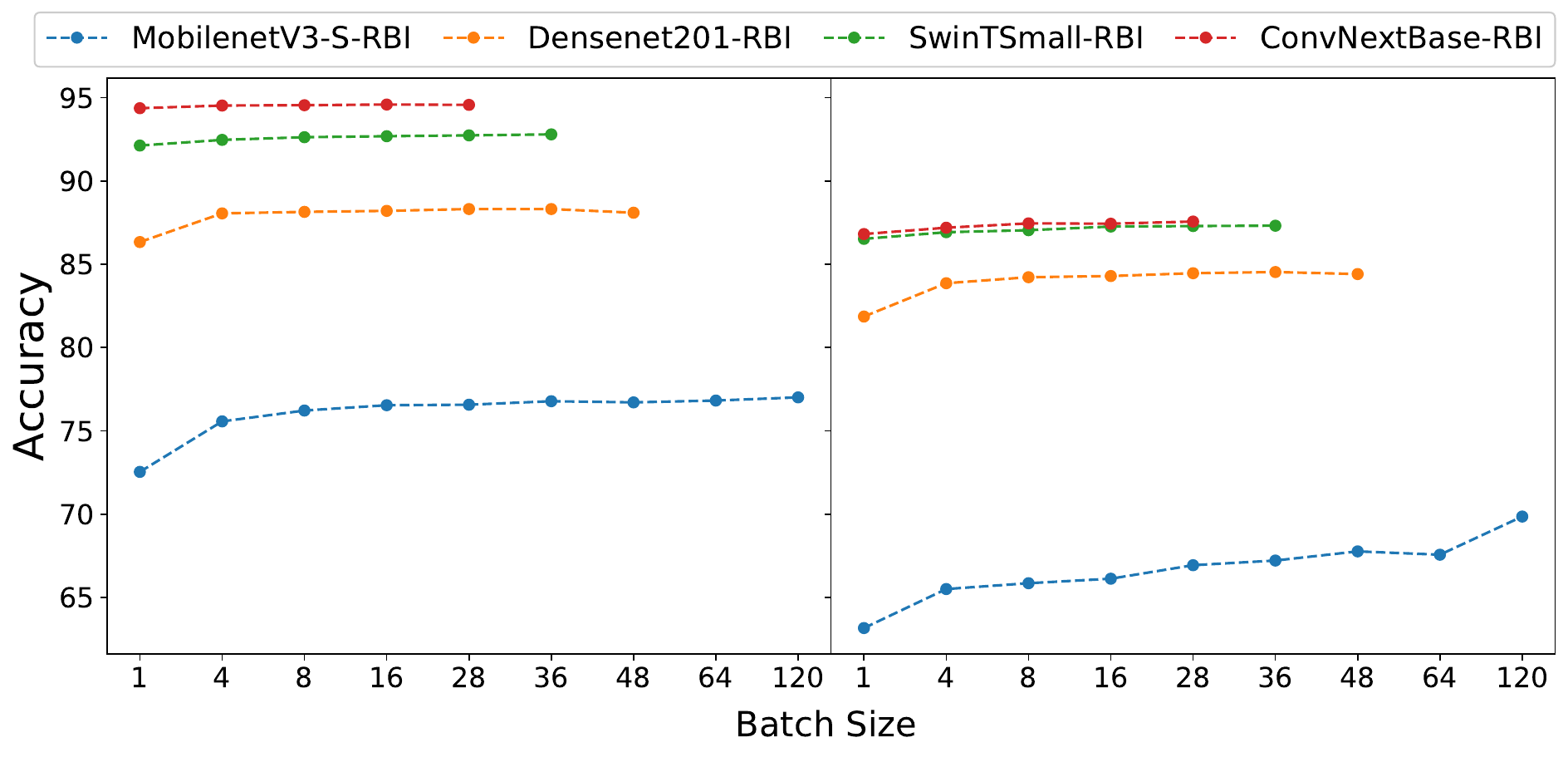}
   \caption{Performance comparison for RBIs using various batch sizes on both the Stanford Dogs dataset (on the left) and the CUB-200-2011 dataset (on the right). Note that experiments with large batch sizes on Densenet201-RBI, SwinT-Small-RBI, and ConvNeXtBase-RBI are omitted due to the GPU's memory constraints.}
   \label{fig:batchsize}
   \vspace{-0.5cm}
\end{figure}

In summary, our proposed approach consistently demonstrates enhanced performance across various classifiers and fine-grained datasets. Moreover, our method can easily integrate with cutting-edge classifiers to yield further enhancements. Notably, the parameter configuration for our approach remains uncomplicated, delivering favorable outcomes with a single setup across diverse classifiers and datasets.

\subsection{The Impact of Batch Configurations}

\label{sec423}
In both the training and inference phases of the proposed module, the feature learning process of the RRA encoder begins by establishing relationships between the features of the DNN encoder within a batch. Therefore, batch configurations, including batch size and how images are selected, influence the model's performance to some extent. In this experiment, we examine the stability of RBI under different batch configurations.

\noindent\textbf{Batch size}. 
Figure \ref{fig:batchsize} demonstrates that varying the batch size during the testing process while keeping the batch size fixed during the training phase has minimal impact on the accuracy of RBI.
Note that in this experiment, we only compare the results of 4 out of the 9 RBI variants for ease of illustration, but other variants exhibit similar trends. The results plotted on both datasets demonstrate that larger models tend to exhibit higher stability, i.e., changes in batch size do not significantly affect performance. From the figure, one can observe that MobilenetV3-S-RBI exhibits the biggest variability. In contrast, the other 3 variants show differences of less than 1\%. It is worth highlighting that despite reducing the batch size to 1, these models are able to maintain their accuracy due to their training methodology, which fully utilizes the available RAM resources.
\subsection{Feature Extracted by Conventional DNN and RBI}

\begin{figure}
\includegraphics[width=\linewidth]{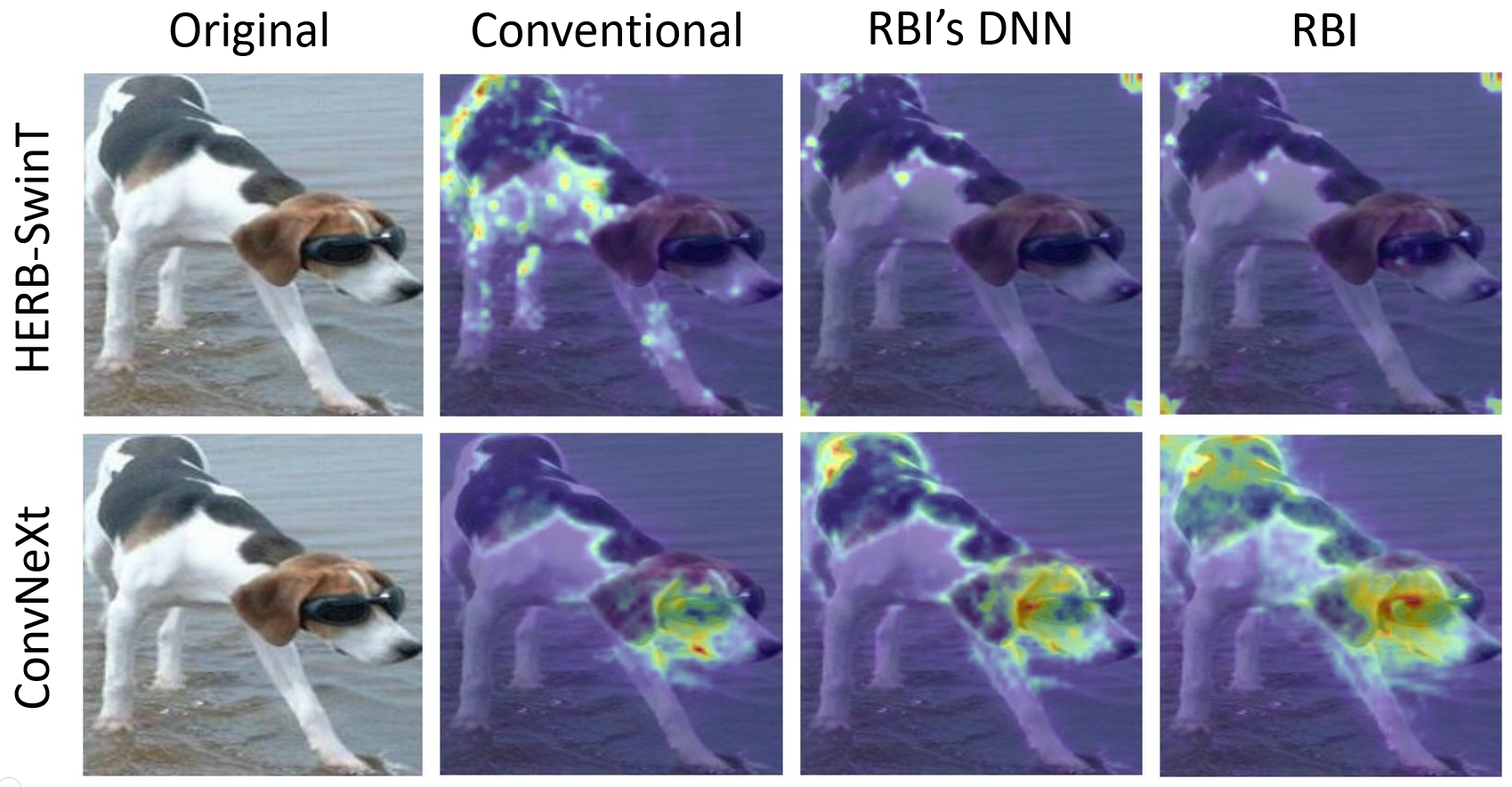}
\caption{Comparison between features extracted by ConvNeXt-Large, ConvNeXt-Large-RBI, HERB-SwinT and HERB-SwinT-RBI on Stanford Dogs dataset, illustrated by GradCam.}
\vspace{-0.5cm}
\label{fig:gph_vs_traditional_gradcam}
\end{figure}

\begin{figure*}
  \centering
   \includegraphics[width=0.8\linewidth]{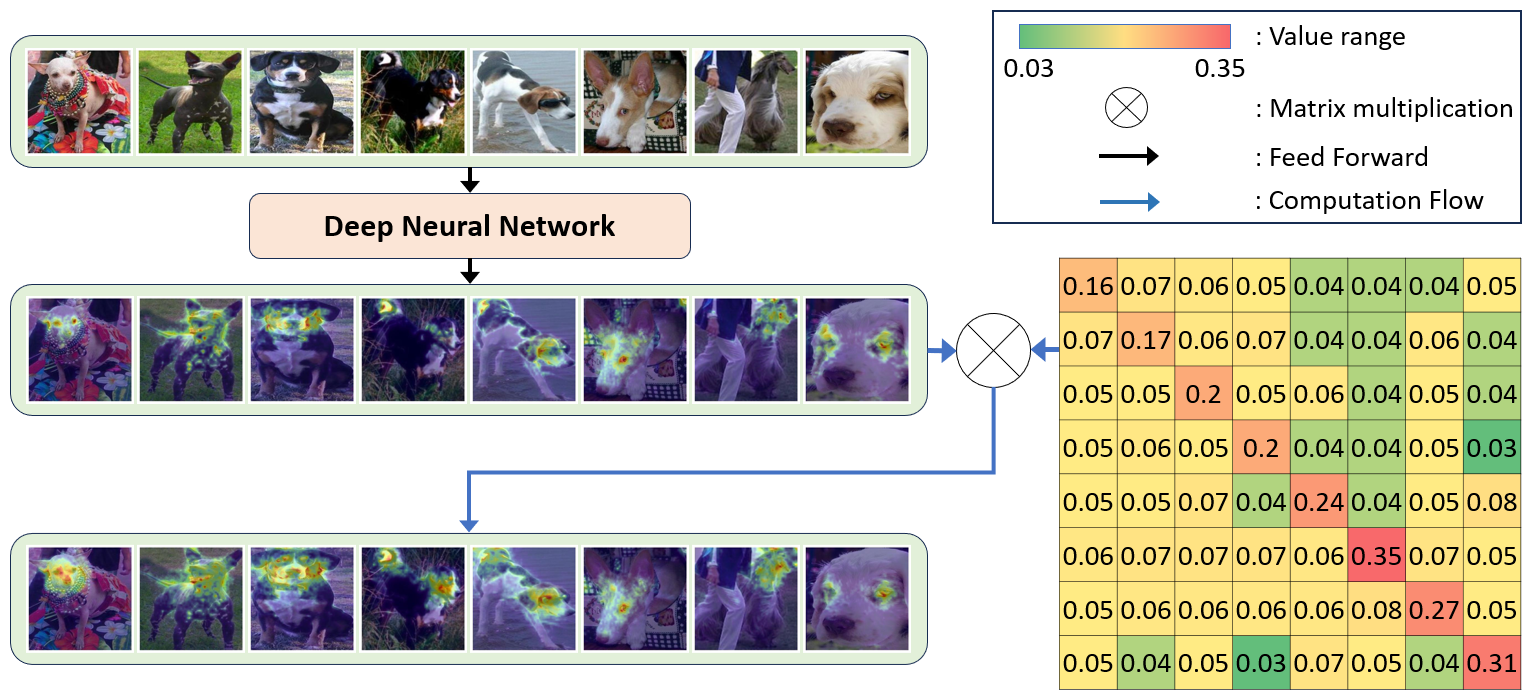}
   \caption{The flow chart illustrates the GradCAM visualizations of features extracted by ConvNeXt-Large-RBI within a batch containing 8 images.}
   \label{fig:gph_insight}
   \vspace{-0.2cm}
\end{figure*}

\begin{figure}[ht]
  \centering
   \includegraphics[width=1\linewidth]{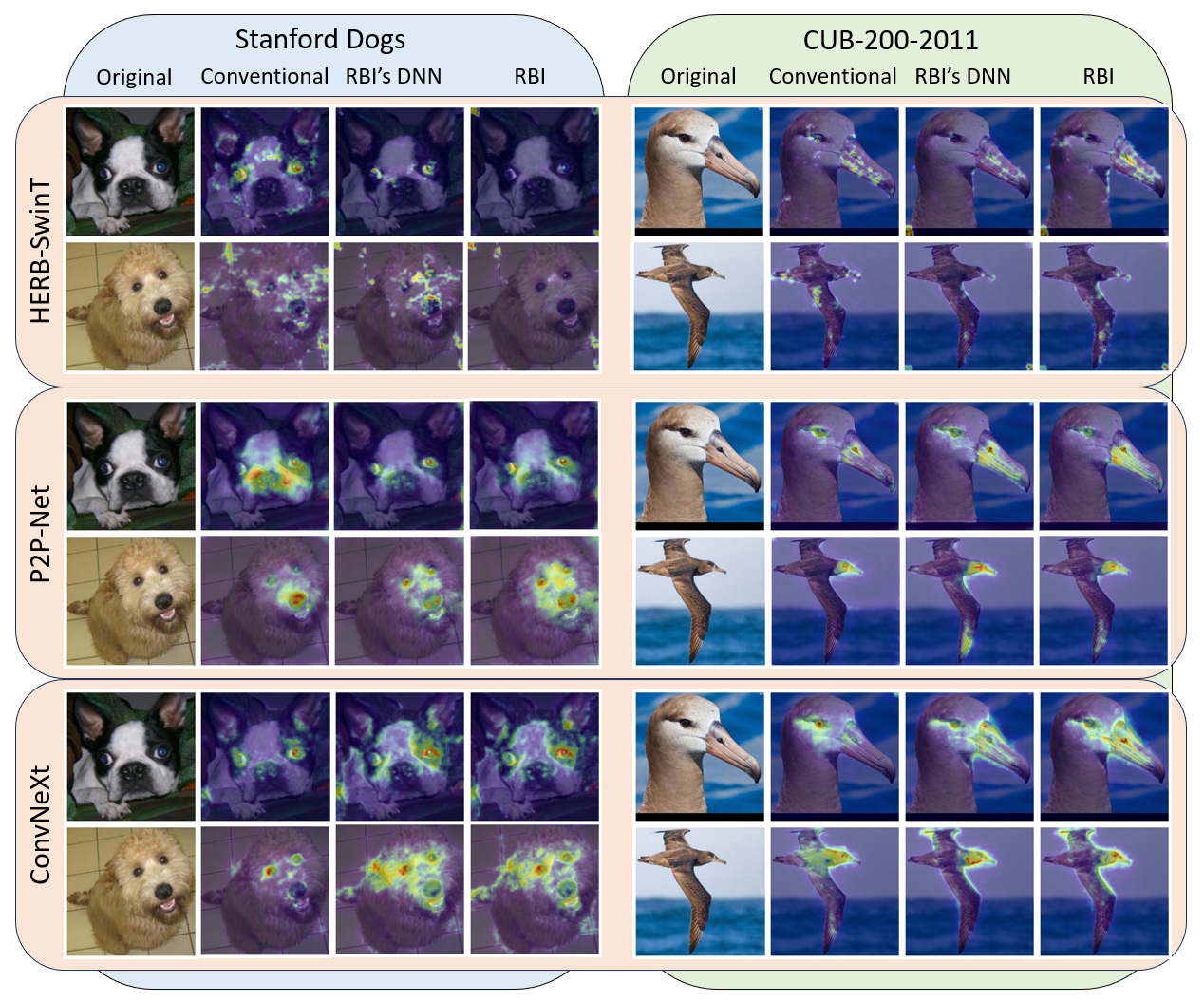}
   \caption{
   Feature visualization comparison using GradCAM: Conventional and RBI models, diverse datasets (Stanford Dogs, CUB-200-2011), and various backbone architectures (HERB-SwinT, P2P-Net, ConvNeXt-Large).}
   \label{fig:gradcam_experiments}
   \vspace{-0.5cm}
\end{figure}

\begin{figure}[ht]
    \centering
    \begin{minipage}{0.23\textwidth}
        \centering
        \includegraphics[width=0.9\textwidth]{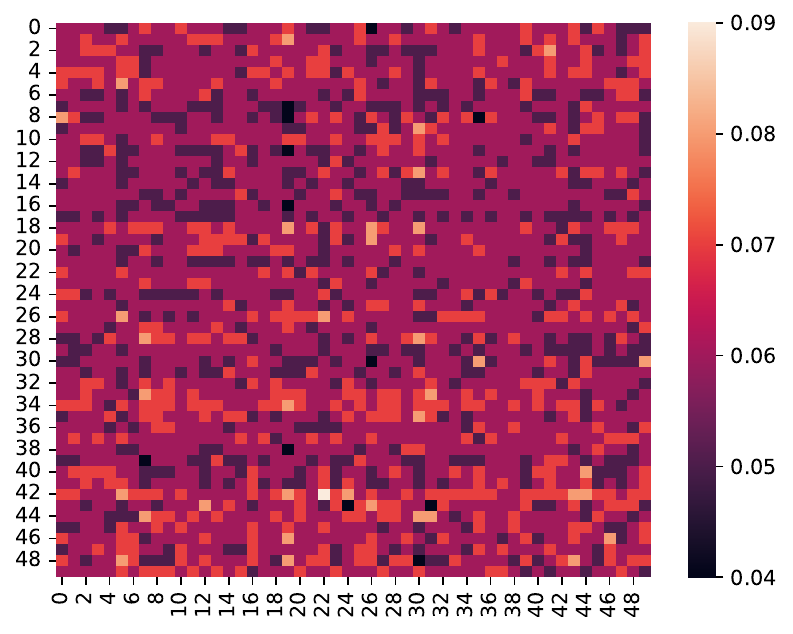}
        \subcaption{}
        \label{fig:gnn_adjacent_matric_heatmap_a}
    \end{minipage}
    \hfill
    \begin{minipage}{0.23\textwidth}
        \centering
        \includegraphics[width=0.9\textwidth]{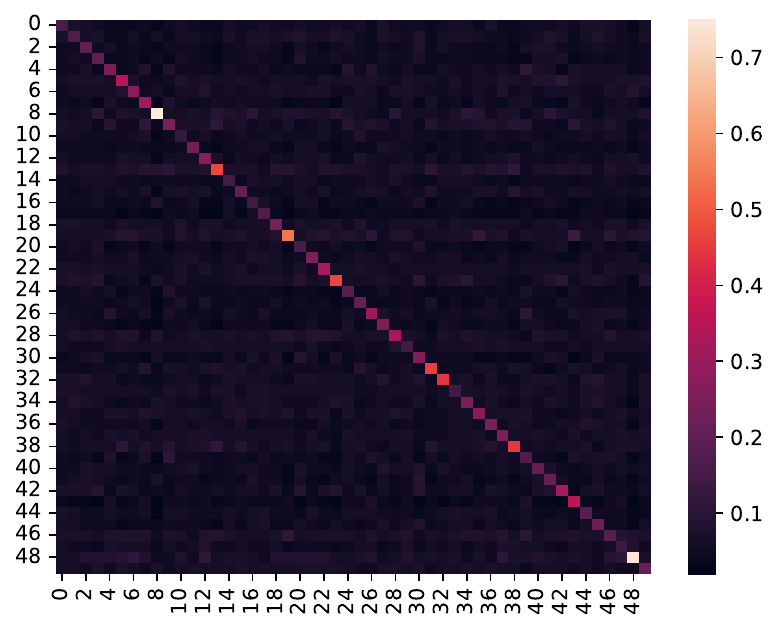}
        \subcaption{}
        \label{fig:gnn_adjacent_matric_heatmap_b}
    \end{minipage}
    \caption{The heatmap of the similarity matrix representing the relationship between images in a batch generated by 
RRA. (a) without RPE, (b) with RPE.}
    \label{fig:gnn_adjacent_matric_heatmap}
    \vspace{-0.6cm}
\end{figure}


Recall that, in our proposed RBI framework, we train the DNN and RRA components simultaneously within an end-to-end framework, enabling the direct influence of RRA backpropagation on DNN parameters during training. As a result, features generated by a conventional DNN backbone exhibit distinct characteristics compared to those produced by the DNN component in the RBI method. Specifically, the RRA tends to propagate gradients backward, allowing the DNN to discern subtler features for combinatorial purposes and create intricate, rich feature maps. Conversely, traditional DNN parameters are propagated to emphasize the most critical feature, assisting in the classification of images into distinct classes.

GradCAM \cite{gradcam} visualizations in Figure \ref{fig:gph_vs_traditional_gradcam} highlight aspects of the aforementioned argument. Here, we employ ConvNeXt-Large, and HERB-SwinT as the DNN backbone. The top row presents GradCAM visualizations for HERB-SwinT, while the bottom row illustrates ConvNeXt visualizations. Moving from left to right in each row, the columns depict features from a traditional method, features extracted by RBI's DNN, and the combined features of both DNN and RRA.


The detailed flow chart in Figure \ref{fig:gph_insight} illustrates the GradCAM visualizations of features extracted by ConvNeXt-Large-RBI, offering a step-by-step insight into the process.  Initially, raw images undergo processing through the DNN backbone, followed by feature structuring into a fully connected graph via the RRA module. Through iterative training, the RRA module learns the relationship between image features, as manifested in the RRA similarity matrix. Subsequently, the RRA integrates embeddings extracted by RBI's DNN through matrix multiplication, ultimately yielding the RBI combined features matrix.

In ConvNeXt, the traditional method predominantly highlights the dog's head in its feature map, showing less emphasis on colored regions elsewhere on the dog's body. Conversely, employing the RBI method's DNN results in a feature map that captures subtle features across various positions, including the head, colored regions, and even the hips. When examining Attention-based models like HERB-SwinT, the RBI method diverges from conventional approaches by focusing on eliminating redundant features and concentrating solely on specific, subtle features crucial for fine-grained image classification tasks. 

Figure \ref{fig:gradcam_experiments} shows a comprehensive comparison of features visualized by GradCAM, encompassing conventional models and the RBI approach across different datasets such as Stanford Dogs and CUB-200-2011, and using various model backbones like HERB-SwinT, P2P-Net, and ConvNeXt-Large.

\subsection{RRA Similarity matrix}

\label{sec:adjacent_matrix}


In this section, we explore two scenarios: RRA with and without RPE. More specifically, the computation of the similarity weight $s_{ij}$ between two images. The experimental result is visualized in Figure \ref{fig:gnn_adjacent_matric_heatmap}, where the x-axis and y-axis enumerate the image indices within a batch, and each cell corresponding to positions on the x and y axes signifies the connection between the two images. The color intensity mirrors the strength of relationships between images, with brighter hues denoting stronger connections and darker shades indicating weaker ones. The matrix observed from Figure \ref{fig:gnn_adjacent_matric_heatmap_a} that, without RPE, lacks intuitiveness, whereas the matrix generated by RRA with RPE in Figure \ref{fig:gnn_adjacent_matric_heatmap_b} is more easily comprehensible. This is evident from the prominent values along the diagonal and the symmetrical appearance of the matrix across the main diagonal. 
There are two benefits to having diagonally symmetrical elements and larger diagonal elements in the similarity matrix. Firstly, it enhances the understanding and intuitiveness of how RRA synthesizes images in batches. Secondly, it leads to a reduced loss of original image information, resulting in a slight improvement in accuracy.\footnote{Despite the quadratic increase in the size of the similarity matrix with batch size, which could potentially lead to memory constraints when dealing with very large batches, this issue is not a significant bottleneck in practical applications. A more detailed discussion on this matter is provided in the Memory Usage section of the supplementary material.} 


\vspace{-0.3cm}
\section{Conclusion}
\label{sec:conclusion}


In this study, we has proposed a novel framework that, while deceptively straightforward, has not been explored in prior works. Empirical experiments on benchmark datasets are conducted to show the effectiveness and to demonstrate seamless integration with various fine-grained classifiers for significant accuracy improvements of our proposed approach. Our architectural innovation also significantly reduces model parameters and inference latency compared to conventional DNN architectures.

Our research opens several promising avenues for future exploration. These include optimizing the architecture and hyperparameters of the integrated RRA-DNN model for different fine-grained classification tasks, exploring different strategies for constructing similarity matrices and RBI frameworks, and applying this approach to other computer vision tasks and datasets to enhance various aspects of visual recognition.

\bibliography{aaai25}

\end{document}